\documentclass[10pt,twocolumn,letterpaper]{article}
\usepackage{iccv}      

\definecolor{iccvblue}{rgb}{0.21,0.49,0.74}
\usepackage[pagebackref,breaklinks,colorlinks,allcolors=iccvblue]{hyperref}
\usepackage{multirow}
\usepackage{stfloats}

\def\paperID{*****} 
\def\confName{ICCV}
\def\confYear{2025}

\title{Your Super Resolution Model is not Enough for Tackling Real-World Scenarios}

\author{DongSik Yoon\\
HDC LABS\\
Seoul, Republic of Korea\\
{\tt\small kevinds1106@hdc-labs.com}
\and
Jongeun Kim\\
HDC LABS\\
Seoul\textbf{}, Republic of Korea\\
{\tt\small JongeunKim@hdc-labs.com}
}

\begin{document}
\maketitle
\begin{abstract}
\label{sec:abstract}
Despite remarkable progress in Single Image Super-Resolution (SISR), traditional models often struggle to generalize across varying scale factors, limiting their real-world applicability. To address this, we propose a plug-in \textbf{Scale-Aware Attention Module (SAAM)} designed to retrofit modern fixed-scale SR models with the ability to perform arbitrary-scale SR. SAAM employs lightweight, scale-adaptive feature extraction and upsampling, incorporating the Simple parameter-free Attention Module (SimAM) for efficient guidance and gradient variance loss to enhance sharpness in image details. Our method integrates seamlessly into multiple state-of-the-art SR backbones (e.g., SCNet, HiT-SR, OverNet), delivering competitive or superior performance across a wide range of integer and non-integer scale factors. Extensive experiments on benchmark datasets demonstrate that our approach enables robust multi-scale upscaling with minimal computational overhead, offering a practical solution for real-world scenarios.
\end{abstract}
\section{Introduction}
\label{sec:intro}
Over the past several years, advances in computer vision have driven significant progress in image super-resolution technology~\cite{sr_survey1, sr_survey2}. 
In particular, Single Image Super-Resolution (SISR) technology has evolved more rapidly and extensively than any other field in computer vision, becoming one of the field’s most widely adopted techniques.
To better accommodate real-world applications, many researchers have continuously investigated ways to balance model efficiency with reconstruction quality.
As a result, recent work has prioritized lightweight architectures while still improving their performance ~\cite{hitsr, lw1, lw2, lw3, lw4, overnet, scnet}.

Furthermore, in applications that receive user input, such as mobile or web-based platforms, upscaling often must produce a fixed output resolution despite variations in the input image size. Traditional super-resolution methods, which rely on fixed scaling factors (e.g., ×2, ×3), struggle with this variability because they require separate models trained for each factor. This leads to increased storage demands and computational overhead, making these approaches impractical for resource-constrained environments in mobile devices and edge-computing platforms.

To overcome these challenges, arbitrary-scale super-resolution has been explored as an alternative. Unlike traditional fixed-scale models, arbitrary-scale SR leverages continuous representations to adapt dynamically to any scaling factor, eliminating the need for multiple pre-trained models. 
By learning representations that generalize beyond discrete training scales, arbitrary-scale SR reduces memory consumption and enhances computational efficiency while maintaining high image quality. Its flexibility to handle varying input resolutions makes it ideal for real-world applications that demand reliable and adaptive upscaling in diverse situations.
For example, ArbSR~\cite{arbsr} introduces a plug-in module that reinforces standard SR backbones with extra scale-aware feature adaptation and upsampling layers.

\begin{figure}[!t]
  \includegraphics[scale=0.525]{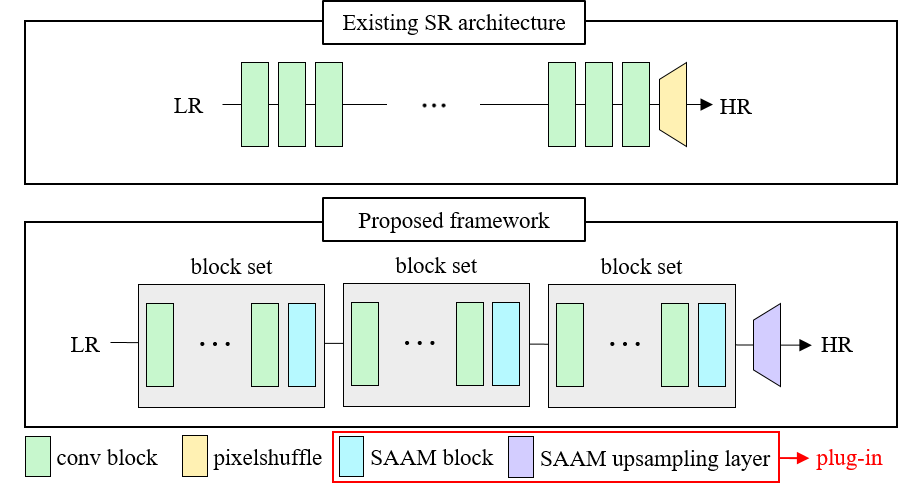}
  \caption{Illustration of the existing SR architecture and the proposed framework incorporating the SAAM plug-in.}
  \label{fig:fig1}
\end{figure}

Despite active research, the integration of arbitrary-scale techniques into modern SR models for real-world deployment remains largely underexplored. Exisitng plug-in modules are often evaluated on outdated or underperforming baselines that are no longer relevant to current practice.
As SR models advance rapidly, it is essential that arbitrary-scale plug-ins are designed to be lightweight, portable, and compatible with cutting-edge models—empowering users to choose solutions tailored to their specific needs.
To address this challenge, we present a novel approach that extends fixed-scale SR models with a plug-in designed for arbitrary scaling. At the core of our method is the Scale-Aware Attention Module \textbf{(SAAM)}—a compact and efficient plug-in that enables flexible scaling with minimal additional parameters, while seamlessly integrating into existing lightweight SR networks.

To mitigate artifacts commonly observed when training on datasets with multi-scale, we incorporate Gradient Variance (GV) loss~\cite{gvloss} into our training objective, promoting sharper, more consistent image reconstructions.  
In addition, recognizing that many earlier arbitrary-scale solutions rely on outdated model backbones, we conduct extensive evaluations of \textbf{SAAM} across a range of state-of-the-art SR architectures to validate its broad applicability and performance.
\section{Related Work}
\label{sec:related}

\subsection{Lightweight Single Image Super-Resolution}

Early CNN-based Super-Resolution (SR) models such as EDSR~\cite{edsr} and RCAN~\cite{rcan} achieve high-fidelity results, but at a considerable computational cost. 
To improve efficiency, real-time networks like FSRCNN \cite{fsrcnn} and ESPCN \cite{espcn} employ lightweight sub-pixel convolutional architectures, demonstrating that processing speed can be increased with only a negligible loss in reconstruction quality.

Extending this line of work, recent works~\cite{imdn, carn, rfdn} explore lightweight designs that preserve performance.
AsConvSR~\cite{lw3} accelerates inference speed by reorganizing convolutional kernels, while MSFIN~\cite{lw1} exploits multi-scale feature interaction to enhance computational efficiency.
The fully-1×1 convolutional networks (SCNet)~\cite{scnet} further reduces the model size by removing conventional 3×3 spatial filters and substituting them with parameter-free spatial shift operations followed by 1×1 convolutions.
Transformer-based models have also emerged: HiT-SR~\cite{hitsr} applies hierarchical self-attention to capture long-range context with minimal memory overhead, and SMFANet~\cite{lw2} aggregates self-modulated features for efficient enhancement of image quality.
Nevertheless, current SR frameworks still target fixed upscaling factors (×2, ×3, ×4), limiting their versatility in real-world applications that require arbitrary resolution scaling.

\subsection{Arbitrary-Scale Super-Resolution}
To overcome the limitations of fixed-scale SR, recent studies have delved into scale-continuous representations that allow a single model to handle arbitrary scale factors.
Among early efforts, ArbSR~\cite{arbsr} introduces a lightweight plug-in module that augments standard SR backbones (e.g., EDSR, RCAN) with scale-aware feature adaptation and upsampling, enabling a single network to handle non-integer and asymmetric magnification factors at minimal cost.
On the other hand, CiaoSR~\cite{arbsr1} proposes a continuous implicit attention-in-attention network that learns scale-aware ensemble weights, fusing local and non-local cues for high-quality arbitrary-scale SR.
To enhance robustness, Scale-Equivariance Pursuit (EQSR)~\cite{arbsr2} designs equivariant processing blocks whose responses transform consistently across scales, yielding more robust feature representations for deep arbitrary-scale SR.
Local Implicit Normalizing Flow (LINF)~\cite{arbsr3} models the conditional distribution of high-resolution textures with a coordinate-conditioned normalizing flow, recovering fine details and mitigating over-smoothing at any requested scale.
Concurrently, OPE-SR~\cite{arbsr4} proposed orthogonal position encoding, a 2D Fourier basis coordinate code, and a parameter free OPE Upscale Module that can be plugged into existing networks for efficient, accurate arbitrary-scale upsampling without additional learnable parameters. 
Although these methods advance performance, most rely on bespoke or outdated backbones, limiting their seamless adoption in the latest compact SR frameworks.


\section{Method}
\label{sec:method}

\begin{figure}[!t]
  \includegraphics[scale=0.425]{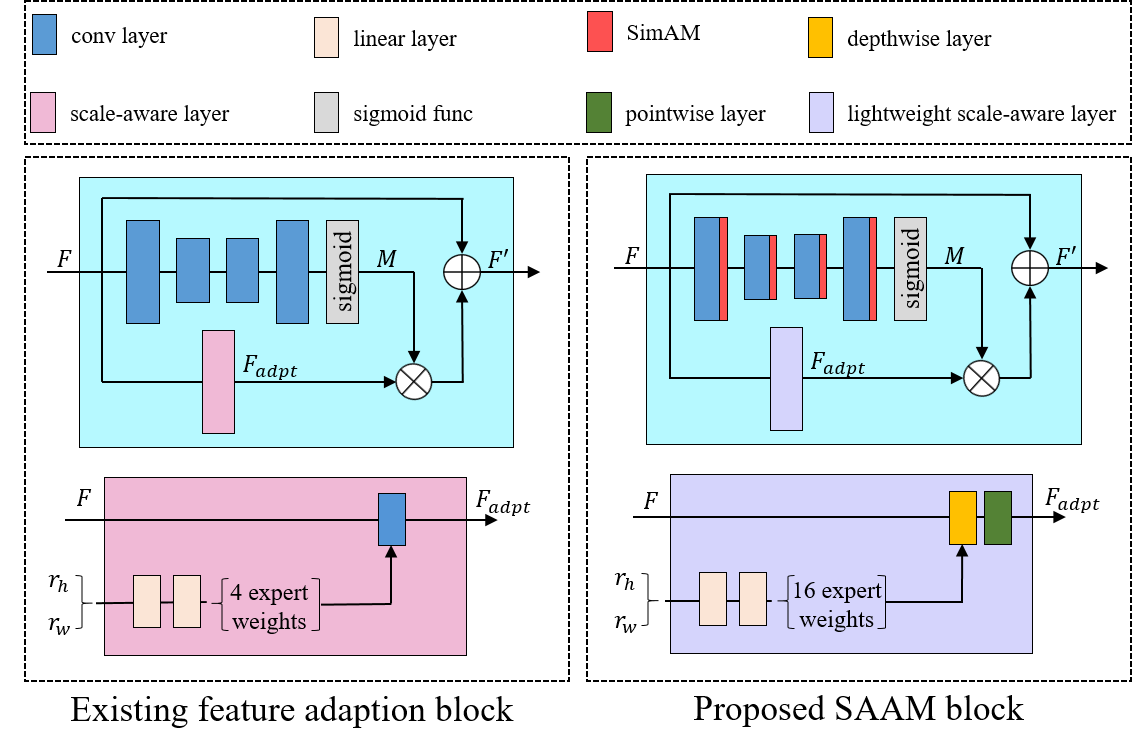}
  \caption{Comparison between the existing feature adaptation block~\cite{arbsr} and the proposed SAAM block.}
  \label{fig:fig2}
\end{figure}

\subsection{Background}
Standard super-resolution (SR) models, as illustrated in Figure~\ref{fig:fig1}, extract feature vectors by stacking multiple convolutional blocks from various existing studies. These extracted features are then passed through an upsampling layer to generate the high-resolution (HR) image. 
In our approach, we employ existing convolutional blocks (e.g., SC‐ResBlock in SCNet~\cite{scnet} and the Residual Hierarchical Transformer Block in HiT‐SR~\cite{hitsr}) without modification, inserting a lightweight scale‐adaptation layer after every $K$ such blocks. Specifically, following each group of $K$ convolutional blocks, we add our proposed plug-in to adapt the feature maps to the desired scale.
Furthermore, inspired by ArbSR~\cite{arbsr}, we replace the conventional upscaling block with a lightweight scale‐aware upsampling block that can handle arbitrary scaling factors. This design enables the network to produce high‐quality HR output at any requested scale.

\subsection{Proposed Framework}
In this work, we propose \textbf{the Scale-Aware Attention Module (SAAM)}, a lightweight yet expressive plug-in that can be seamlessly integrated into existing SR networks.
SAAM allows the network to adapt dynamically to the target scaling factor, including non-integer and asymmetric scales, without a prohibitive increase in parameters.

\subsubsection*{SAAM Block}
Figure~\ref{fig:fig2} contrasts the existing feature-adaptation block with our proposed SAAM block. Following ArbSR(left), the feature map $F$ produced by the preceding convolutional stage is first fed into an hourglass module to generate a guidance map $M$. To further refine $M$, we replace batch normalization inside the hourglass with SimAM~\cite{simam}, a parameter-free attention mechanism that assigns full three-dimensional (spatial + channel) weights. This substitution preserves the lightweight footprint of the hourglass while more precisely emphasizing salient local structures and suppressing irrelevant regions, resulting in a highly discriminative guidance map.
In parallel, $F$ is processed by a lightweight scale-aware convolution to obtain the adapted features $F_{\mathrm{adpt}}$. We then fuse the three signals as shown in Equation \ref{eq:1},

\begin{equation}
F' = F + F_{adpt} \times M,
\label{eq:1}
\end{equation}

where the guidance map $M$ acts as a gate, modulating the contribution of $F_{\mathrm{adpt}}$ relative to the original $F$ according to the similarity of local features.

\subsubsection*{Lightweight Scale-Aware Layer}
Existing scale-aware convolutional layers in ArbSR learn multiple groups of expert kernels, each capturing distinct scale-specific patterns. Given horizontal and vertical magnification factors ($r_{h}$, $r_{v}$ in Figure~\ref{fig:fig2}), the model dynamically blends these experts through learned routing weights to synthesize the final filters. Conditioning on scale information in this way enables a single network to accommodate a wide range of magnification factors without requiring a separate model for each scale. Although increasing the number of experts can improve performance by providing finer scale control, it also inflates the parameter count and overall model complexity.

To achieve a better trade-off, we introduce a lightweight scale-aware convolutional layer (Figure~\ref{fig:fig2}, right)  that preserves scale adaptability while substantially reducing parameter overhead.
 Instead of enlarging the expert set, we factorize the dynamic convolution into a depthwise (group-wise) operation and followed by a point-wise convolution, thereby decoupling the channel dimension from the dynamic filter generation. Concretely, the fused expert kernel is applied only to groups of channels, and then the resulting features are merged through the 1 × 1 convolution.
This design retains the ability to model scale-specific representations while curtailing parameter growth. 
As a result, our SAAM block efficiently handles diverse magnification factors while maintaining a lightweight, scalable structure.

\subsubsection*{SAAM Upsampling Layer}
We lighten the computational load of the plug-in by redesigning ArbSR’s scale-aware upsampling layer.

First, the standard convolution is replaced by a depth-wise convolution, reducing the number of parameters.
Second, we replace batch normalization with the SimAM module, which enhances feature representations without introducing any extra trainable parameters.
Lastly, informed by studies demonstrating that Swish (SiLU) outperforms ReLU in super-resolution tasks \cite{silu1, silu2}, we substitute SiLU for every ReLU activation in the upsampling path.
As a result, these changes yield a lightweight SAAM upsampling layer that preserves the reconstruction quality of the original design while markedly lowering model complexity.

\subsection{Loss Function}
To preserve fine edges while keeping the model compact, we combine the conventional $L_{1}$ loss with the gradient variance (GV) loss \cite{gvloss}.
GV loss begins by applying the Sobel operators to grayscale versions of the ground-truth image $I^{HR}$ and the generated image $I^{SR}$, producing gradient maps $G^{HR}_{x}$, $G^{HR}_{y}$, $G^{SR}_{x}$, and $G^{SR}_{y}$. 
The variance maps $V^{HR}_{x}$, $V^{HR}_{y}$, $V^{SR}_{x}$, and $V^{SR}_{y}$ are then calculated from these gradients. Because generated images typically exhibit smoother gradients, their local variances are lower than those of $I^{HR}$. Capitalizing on this observation, the GV loss is defined as Equation \ref{eq:2}:

\begin{gather}
 L_{GV}=\mathbb{E}\|V^{HR}_{x} - V^{SR}_{x}\|_{2} + \mathbb{E}\|V^{HR}_{y} - V^{SR}_{y}\|_{2}
 \label{eq:2}
\end{gather}

Minimizing $L_{\text{GV}}$ narrows the variance gap and promotes sharper and clearer edges. The full training objective is therefore given by Equation \ref{eq:3}

\begin{gather}
 L_{all} = \|I^{HR} - I^{SR}\|_{1} + \lambda_{gv}L_{GV}
 \label{eq:3}
\end{gather}

where $\lambda$ balances edge fidelity against overall reconstruction accuracy.

\section{Experiments}
\label{sec:experiments}


\begin{table*}[!t]
\centering
\resizebox{\textwidth}{!}{
\begin{tabular}{|lccccccccc}
\hline
\multicolumn{1}{|l||}{Scale}                                                             & \multicolumn{3}{c||}{×2}                                                                                         & \multicolumn{3}{c||}{×3}                                                                                               & \multicolumn{3}{c||}{×4}                                                                                               \\ \hline
\multicolumn{1}{|l||}{Metrics}                                                           & \multicolumn{3}{c||}{PSNR / SSIM}                                                                                & \multicolumn{3}{c||}{PSNR / SSIM}                                                                                      & \multicolumn{3}{c||}{PSNR / SSIM}                                                                                      \\ \hline
\multicolumn{1}{|l||}{Dataset}                                                           & \multicolumn{1}{c|}{BSD100}           & \multicolumn{1}{c|}{Urban100}           & \multicolumn{1}{c||}{Manga109}           & \multicolumn{1}{c|}{BSD100}             & \multicolumn{1}{c|}{Urban100}             & \multicolumn{1}{c||}{Manga109}             & \multicolumn{1}{c|}{BSD100}             & \multicolumn{1}{c|}{Urban100}             & \multicolumn{1}{c||}{Manga109}             \\ \hline\hline
\multicolumn{1}{|l||}{\begin{tabular}[c]{@{}l@{}}SCNet ×2 \\ (146K)\end{tabular}}        & \multicolumn{1}{c|}{31.92 / 0.8963} & \multicolumn{1}{c|}{31.14 / 0.9175} & \multicolumn{1}{c||}{37.77 / 0.9752} & \multicolumn{1}{c|}{-}                & \multicolumn{1}{c|}{-}                & \multicolumn{1}{c||}{-}                & \multicolumn{1}{c|}{-}                & \multicolumn{1}{c|}{-}                & \multicolumn{1}{c||}{-}                \\ \hline
\multicolumn{1}{|l||}{\begin{tabular}[c]{@{}l@{}}SCNet ×3 \\ (146K)\end{tabular}}        & \multicolumn{1}{c|}{-}              & \multicolumn{1}{c|}{-}              & \multicolumn{1}{c||}{-}              & \multicolumn{1}{c|}{28.86 / 0.7988} & \multicolumn{1}{c|}{27.38 / 0.8344} & \multicolumn{1}{c||}{32.50 / 0.9374} & \multicolumn{1}{c|}{-}                & \multicolumn{1}{c|}{-}                & \multicolumn{1}{c||}{-}                \\ \hline
\multicolumn{1}{|l||}{\begin{tabular}[c]{@{}l@{}}SCNet ×4 \\ (154K)\end{tabular}}        & \multicolumn{1}{c|}{-}              & \multicolumn{1}{c|}{-}              & \multicolumn{1}{c||}{-}              & \multicolumn{1}{c|}{-}                & \multicolumn{1}{c|}{-}                & \multicolumn{1}{c||}{-}                & \multicolumn{1}{c|}{27.37 / 0.7293} & \multicolumn{1}{c|}{25.49 / 0.7648} & \multicolumn{1}{c||}{29.52 / 0.8959} \\ \hline
\multicolumn{1}{|l||}{\begin{tabular}[c]{@{}l@{}}SCNet+Ours(T)\\ (228K)\end{tabular}}    & \multicolumn{1}{l|}{\textbf{31.99 / 0.8977}} & \multicolumn{1}{l|}{\textbf{31.43 / 0.9215}} & \multicolumn{1}{l||}{\textbf{38.07 / 0.9756}} & \multicolumn{1}{l|}{\textbf{28.89} / 0.7977}   & \multicolumn{1}{l|}{\textbf{27.53 / 0.8371}}   & \multicolumn{1}{l||}{\textbf{32.62} / 0.9372}   & \multicolumn{1}{l|}{27.15 / 0.7139}   & \multicolumn{1}{l|}{25.14 / 0.7485}   & \multicolumn{1}{l||}{28.64 / 0.8815}   \\ \hline
\multicolumn{1}{|l||}{\begin{tabular}[c]{@{}l@{}}SCNet+Ours(L)\\ (505K)\end{tabular}}   & \multicolumn{1}{l|}{\textbf{31.94 / 0.8985}} & \multicolumn{1}{l|}{\textbf{31.30 / 0.9212}} & \multicolumn{1}{l||}{\textbf{37.86 / 0.9756}} & \multicolumn{1}{l|}{\textbf{28.94 / 0.8026}}   & \multicolumn{1}{l|}{\textbf{27.61 / 0.8412}}   & \multicolumn{1}{l||}{\textbf{32.97 / 0.9410}}   & \multicolumn{1}{l|}{27.35 / 0.7230}   & \multicolumn{1}{l|}{25.45 / 0.7607}   & \multicolumn{1}{l||}{29.49 / 0.8939}   \\ \hline\hline
\multicolumn{10}{|l|}{}                                                                                                                                                                                                                                                                                                                                                                                                                                   \\ \hline
\multicolumn{1}{|l||}{\begin{tabular}[c]{@{}l@{}}HiT-SIR ×2 \\ (772K)\end{tabular}}      & \multicolumn{1}{c|}{32.35/0.9019}   & \multicolumn{1}{c|}{33.02/0.9365}   & \multicolumn{1}{c||}{39.38/0.9782}   & \multicolumn{1}{c|}{-}                & \multicolumn{1}{c|}{-}                & \multicolumn{1}{c||}{-}                & \multicolumn{1}{c|}{-}                & \multicolumn{1}{c|}{-}                & \multicolumn{1}{c||}{-}                \\ \hline
\multicolumn{1}{|l||}{\begin{tabular}[c]{@{}l@{}}HiT-SIR ×3\\ (772K)\end{tabular}}       & \multicolumn{1}{c|}{-}              & \multicolumn{1}{c|}{-}              & \multicolumn{1}{c||}{-}              & \multicolumn{1}{c|}{29.27 / 0.8101}   & \multicolumn{1}{c|}{28.93 / 0.8673}   & \multicolumn{1}{c||}{34.40 / 0.9496}   & \multicolumn{1}{c|}{-}                & \multicolumn{1}{c|}{-}                & \multicolumn{1}{c||}{-}                \\ \hline
\multicolumn{1}{|l||}{\begin{tabular}[c]{@{}l@{}}HiT-SIR ×4\\ (772K)\end{tabular}}       & \multicolumn{1}{c|}{-}              & \multicolumn{1}{c|}{-}              & \multicolumn{1}{c||}{-}              & \multicolumn{1}{c|}{-}                & \multicolumn{1}{c|}{-}                & \multicolumn{1}{c||}{-}                & \multicolumn{1}{c|}{27.73 / 0.7424}   & \multicolumn{1}{c|}{26.71 / 0.8045}   & \multicolumn{1}{c||}{31.23 / 0.9176}   \\ \hline
\multicolumn{1}{|l||}{\begin{tabular}[c]{@{}l@{}}HiT-SIR+Ours(T)\\ (903K)\end{tabular}}  & \multicolumn{1}{c|}{32.34 / \textbf{0.9021}} & \multicolumn{1}{c|}{\textbf{33.02 / 0.9365}} & \multicolumn{1}{l||}{39.23 / 0.9778} & \multicolumn{1}{c|}{\textbf{29.27 / 0.8111}}   & \multicolumn{1}{c|}{28.89 / \textbf{0.8679}}   & \multicolumn{1}{l||}{34.36 / \textbf{0.9496}}   & \multicolumn{1}{c|}{\textbf{27.76 / 0.7442}}   & \multicolumn{1}{c|}{\textbf{26.75 / 0.8075}}   & \multicolumn{1}{l||}{31.22 / \textbf{0.9180}}   \\ \hline
\multicolumn{1}{|l||}{\begin{tabular}[c]{@{}l@{}}HiT-SIR+Ours(L)\\ (1,385K)\end{tabular}} & \multicolumn{1}{c|}{\textbf{32.35 / 0.9020}} & \multicolumn{1}{l|}{\textbf{33.04 / 0.9369}} & \multicolumn{1}{l||}{39.32 / \textbf{0.9783}} & \multicolumn{1}{c|}{\textbf{29.30 / 0.8113}}   & \multicolumn{1}{l|}{\textbf{28.95 / 0.8688}}   & \multicolumn{1}{l||}{34.39 / \textbf{0.9497}}   & \multicolumn{1}{c|}{\textbf{27.77 / 0.7444}}   & \multicolumn{1}{l|}{\textbf{26.78 / 0.8079}}   & \multicolumn{1}{l||}{\textbf{31.27 / 0.9182}}   \\ \hline\hline
\multicolumn{10}{|l}{}                                                                                                                                                                                                                                                                                                                                                                                                                                    \\ \hline
\multicolumn{1}{|l||}{\begin{tabular}[c]{@{}l@{}}OverNet ×2 \\ (943K)\end{tabular}}      & \multicolumn{1}{c|}{32.24 / 0.8999} & \multicolumn{1}{c|}{32.44 / 0.9304} & \multicolumn{1}{c||}{38.97 / 0.9773} & \multicolumn{1}{c|}{28.12 / 0.7749}   & \multicolumn{1}{c|}{26.35 / 0.8040}   & \multicolumn{1}{c||}{30.12 / 0.9030}   & \multicolumn{1}{c|}{26.78 / 0.7001}   & \multicolumn{1}{c|}{24.59 / 0.7208}   & \multicolumn{1}{c||}{27.48 / 0.8444}   \\ \hline
\multicolumn{1}{|l||}{\begin{tabular}[c]{@{}l@{}}OverNet ×3 \\ (943K)\end{tabular}}      & \multicolumn{1}{c|}{31.95 / 0.8973} & \multicolumn{1}{c|}{31.50 / 0.9242} & \multicolumn{1}{c||}{37.63 / 0.9747} & \multicolumn{1}{c|}{29.11 / 0.8053}   & \multicolumn{1}{c|}{28.29 / 0.8541}   & \multicolumn{1}{c||}{33.72 / 0.9454}   & \multicolumn{1}{c|}{27.51 / 0.7323}   & \multicolumn{1}{c|}{25.96 / 0.7796}   & \multicolumn{1}{c||}{30.16 / 0.9032}   \\ \hline
\multicolumn{1}{|l||}{\begin{tabular}[c]{@{}l@{}}OverNet ×4 \\ (943K)\end{tabular}}      & \multicolumn{1}{c|}{31.34 / 0.8874} & \multicolumn{1}{c|}{29.99 / 0.9072} & \multicolumn{1}{c||}{35.13 / 0.9669} & \multicolumn{1}{c|}{28.90 / 0.7998}   & \multicolumn{1}{c|}{27.75 / 0.8461}   & \multicolumn{1}{c||}{32.61 / 0.9390}   & \multicolumn{1}{c|}{27.59 / 0.7373}   & \multicolumn{1}{c|}{26.18 / 0.7879}   & \multicolumn{1}{c||}{30.56 / 0.9097}   \\ \hline
\multicolumn{1}{|l||}{\begin{tabular}[c]{@{}l@{}}OverNet+Ours(T)\\ (1,076K)\end{tabular}} & \multicolumn{1}{c|}{32.19 / \textbf{0.9001}} & \multicolumn{1}{c|}{32.31 / 0.9296} & \multicolumn{1}{c||}{38.76 / 0.9772} & \multicolumn{1}{c|}{\textbf{29.11 / 0.8066}}   & \multicolumn{1}{c|}{28.27 / \textbf{0.8551}}   & \multicolumn{1}{c||}{\textbf{33.79 / 0.9458}}   & \multicolumn{1}{c|}{\textbf{27.60 / 0.7382}}   & \multicolumn{1}{c|}{26.16 / \textbf{0.7895}}   & \multicolumn{1}{c||}{\textbf{30.66 / 0.9108}}   \\ \hline
\multicolumn{1}{|l||}{\begin{tabular}[c]{@{}l@{}}OverNet+Ours(L)\\ (2,312K)\end{tabular}}  & \multicolumn{1}{c|}{\textbf{32.24 / 0.9006}} & \multicolumn{1}{c|}{\textbf{32.48 / 0.9314}} & \multicolumn{1}{c||}{38.96 / \textbf{0.9777}} & \multicolumn{1}{c|}{\textbf{29.15 / 0.8077}}   & \multicolumn{1}{c|}{\textbf{28.41 / 0.8589}}   & \multicolumn{1}{c||}{\textbf{33.89 / 0.9468}}   & \multicolumn{1}{c|}{\textbf{27.62 / 0.7396}}   & \multicolumn{1}{c|}{\textbf{26.28 / 0.7949}}   & \multicolumn{1}{c||}{\textbf{30.75 / 0.9131}}   \\ \hline\hline
\end{tabular}
}
\caption{Quantitative comparisons by integrating our method into existing models~\cite{scnet, overnet, hitsr}. \textbf{Bold} indicates superior performance over baseline.}
\label{table1}
\end{table*}
We evaluate the proposed method with three state-of-the-art super-resolution (SR) backbones:
\begin{itemize}
    \item SCNet-tiny~\cite{scnet} – a compact CNN to test the effectiveness of the plug-in on ultra-lightweight architectures.
    \item HiT-SIR~\cite{hitsr} – a variant of SwinIR-Light~\cite{swinir}(HiT-SIR), is adopted to confirm compatibility with Transformer-based self-attention models. 
    \item OverNet~\cite{overnet} – a model inherently built for flexible scaling to assess compatibility with arbitrary-scale SR frameworks.
\end{itemize}
In each backbone model, we empirically insert the SAAM block at the point that added the fewest parameters while retaining accuracy.

\subsection{Implement Details}
All experiments are conducted in PyTorch. We set the gradient-variance loss weight
 $\lambda_{gv}$ to 0.01.
In our study, all models are trained on the DIV2K dataset~\cite{div2k}. When pretrained weights are unavailable or have been trained on additional data, we retrain the model from scratch using only the DIV2K dataset. 
Unlike existing works that train a separate network for each upscaling factor, we train on ×2, ×3, and ×4 scales simultaneously.
We report performance on BSD100~\cite{b100}, Urban100~\cite{u100}, and Manga109~\cite{m109}, reporting peak signal-to-noise ratio (PSNR) and structural similarity index (SSIM), the two standard metrics in image enhancement research.

\begin{table*}[!t]
\centering
\resizebox{\textwidth}{!}{
\begin{tabular}{|l|c|cccccc|}
\hline
\multirow{2}{*}{Method} & \multirow{2}{*}{params}   & \multicolumn{6}{c|}{BSD100}                                                                                                                                                                                  \\ \cline{3-8} 
                        &                              & \multicolumn{1}{c|}{×1.2}           & \multicolumn{1}{c|}{×1.6}           & \multicolumn{1}{c|}{×2.4}           & \multicolumn{1}{c|}{×2.8}           & \multicolumn{1}{c|}{×3.2}           & ×3.6           \\ \hline
OverNet~\cite{overnet}                 & \multicolumn{1}{c|}{943K $\ast$ 3} & \multicolumn{1}{c|}{32.30 / 0.9347} & \multicolumn{1}{c|}{32.52 / 0.9226} & \multicolumn{1}{c|}{29.86 / 0.8454} & \multicolumn{1}{c|}{27.49 / 0.7862} & \multicolumn{1}{c|}{\textbf{28.71} / 0.7886} & 27.97 / 0.7605 \\ \hline
Ours                    & \multicolumn{1}{c|}{1,076K}  & \multicolumn{1}{c|}{\textbf{37.98 / 0.9763}} & \multicolumn{1}{c|}{\textbf{34.09 / 0.9388}} & \multicolumn{1}{c|}{\textbf{30.39 / 0.8578}} & \multicolumn{1}{c|}{\textbf{29.39 / 0.8229}} & \multicolumn{1}{c|}{28.64 / \textbf{0.7892}} & \textbf{27.99 / 0.7630} \\ \hline\hline
\multirow{2}{*}{Method} & \multirow{2}{*}{param.}   & \multicolumn{6}{c|}{Manga109}                                                                                                                                                                                \\ \cline{3-8} 
                        &                              & \multicolumn{1}{c|}{×1.2}           & \multicolumn{1}{c|}{×1.6}           & \multicolumn{1}{c|}{×2.4}           & \multicolumn{1}{c|}{×2.8}           & \multicolumn{1}{c|}{×3.2}           & ×3.6           \\ \hline
OverNet~\cite{overnet}                 & \multicolumn{1}{c|}{943K $\ast$ 3} & \multicolumn{1}{c|}{32.15 / 0.9610} & \multicolumn{1}{c|}{35.24 / 0.9719} & \multicolumn{1}{c|}{30.14 / 0.9294} & \multicolumn{1}{c|}{29.97 / 0.9299} & \multicolumn{1}{c|}{31.03 / 0.9286} & 30.87 / 0.9183 \\ \hline
Ours                    & \multicolumn{1}{c|}{1,076K}  & \multicolumn{1}{c|}{\textbf{43.21 / 0.9938}} & \multicolumn{1}{c|}{\textbf{39.73 / 0.9856}} & \multicolumn{1}{c|}{\textbf{35.17 / 0.9611}} & \multicolumn{1}{c|}{\textbf{33.66 / 0.9481}} & \multicolumn{1}{c|}{\textbf{32.34 / 0.9346}} & \textbf{31.15 / 0.9202} \\ \hline
\end{tabular}
}
\caption{Quantitative evaluation of non-integer factor SR via PSNR and SSIM metrics. \textbf{Bold} indicates the best.}
\label{table2}
\end{table*}

\begin{table*}[!t]
\centering
\resizebox{\textwidth}{!}{
\begin{tabular}{|l|c|ccc|ccc|}
\hline
\multirow{2}{*}{{\begin{tabular}[c]{@{}l@{}}Method\\ (HiT-SIR~\cite{hitsr})\end{tabular}}}    & \multirow{2}{*}{params} & \multicolumn{3}{c||}{SAAM Block}                                                 & \multicolumn{1}{c|}{×2}             & \multicolumn{1}{c|}{×3}             & ×4                                  \\ \cline{3-8}
  &           & \multicolumn{1}{c|}{Normalization} & \multicolumn{1}{c|}{Experts} & \multicolumn{1}{c||}{Dense Layer} & \multicolumn{3}{c|}{PSNR / SSIM}                                                                                \\ \hline
Baseline & 772K $\ast$ 3   & \multicolumn{1}{c|}{-}       & \multicolumn{1}{c|}{-}             & \multicolumn{1}{c||}{-}           & \multicolumn{1}{l|}{\underline{32.35} / 0.9019} & \multicolumn{1}{l|}{\underline{29.27} / 0.8101} & \multicolumn{1}{l|}{27.73 / 0.7424} \\ \hline
BN-4        & 872K      & \multicolumn{1}{c|}{Batch}       & \multicolumn{1}{c|}{4}         & \multicolumn{1}{c||}{O}           & \multicolumn{1}{c|}{32.33 / 0.9020} & \multicolumn{1}{c|}{29.26 / \underline{0.8109}} & 27.74 / 0.7438                      \\ \hline
SA-4        & 871K      & \multicolumn{1}{c|}{SimAM}       & \multicolumn{1}{c|}{4}         & \multicolumn{1}{c||}{O}           & \multicolumn{1}{c|}{32.34 / \underline{0.9021}} & \multicolumn{1}{c|}{\textbf{29.28} / \underline{0.8109}} & \underline{27.75} / 0.7438                      \\ \hline
SA-16 (Ours)     & 903K      & \multicolumn{1}{c|}{SimAM}      & \multicolumn{1}{c|}{16}         & \multicolumn{1}{c||}{O}           & \multicolumn{1}{c|}{32.34 / \underline{0.9021}} & \multicolumn{1}{c|}{\underline{29.27} / \textbf{0.8111}} & \textbf{27.76} / \underline{0.7442}                      \\ \hline
SA-16 (no-dense)        & 996K      & \multicolumn{1}{c|}{SimAM}      & \multicolumn{1}{c|}{16}         & \multicolumn{1}{c||}{$\times$}           & \multicolumn{1}{c|}{\textbf{32.36} / \textbf{0.9022}} & \multicolumn{1}{c|}{\textbf{29.28} / \textbf{0.8111}} & \textbf{27.76} / \textbf{0.7443}                      \\ \hline
SA-64        & 1,107K    & \multicolumn{1}{c|}{SimAM}      & \multicolumn{1}{c|}{64}         & \multicolumn{1}{c||}{O}           & \multicolumn{1}{c|}{32.34 / 0.9019} & \multicolumn{1}{c|}{\underline{29.27} / 0.8107} & \underline{27.75} / 0.7436                      \\ \hline
\end{tabular}
}
\caption{Ablation study performed by our network with different settings BSD100~\cite{b100}. \textbf{Bold} and \underline{Underline} indicate the best and second, respectively.}
\label{table3}
\end{table*}

\subsection{Quantitative Comparisons}
We evaluate our approach under two configurations. 
The large version (L) mirrors the original ArbSR plug-in \cite{arbsr} and omits SimAM and depth-wise layers, whereas the tiny version (T) adopts the lightweight SAAM design.

Table~\ref{table1} shows that baseline models \cite{scnet,hitsr,overnet} are each trained on data that match a single output scale; as a result, most cannot infer images at unseen scales. Although they reach state-of-the-art accuracy on their respective scales, they inherently lack the flexibility required for wide-range upscaling. 
By contrast, our unified model, with only a lightweight plug-in, delivers comparable or superior results across many scales. In particular, HiT-SIR+Our(L) outperforms the original single-scale HiT-SIR on nearly every dataset and scale factor. The HiT-SIR+Our(T) variant provides strong multi-scale capability while adding merely 1.2 K parameters.
Although OverNet natively supports multi-scale inference, our method surpasses it on every benchmark dataset and scale, highlighting the plug-in’s effectiveness. For SCNet, with its exceptionally small parameter budget, our plug-in exceeds accuracy on BSD100 and Manga109 at ×2 and ×3, though performance declines at ×4; we attribute this drop to the backbone’s limited capacity.

To verify non-integer upscaling, we conduct further tests on BSD100 and Manga109 using the OverNet backbone. We run the OverNet model pretrained at ×2 to test on scales on ×1.2, ×1.6, and ×2.4; the model pretrained at ×3 for ×2.8 and ×3.2; and the model pretrained at ×4 for ×3.6. In Table~\ref{table2}, the symbol $\ast$ marks results that require multiple pretrained networks. Despite using a single set of weights, our method delivers higher PSNR and SSIM at these fractional scales while consuming far fewer parameters and resources.

Consequently, we demonstrate that integrating the proposed SAAM plug-in into the latest SR architectures not only enhances their performance but also enables arbitrary-scale SR within a single model.

\subsection{Ablation Study}
To examine the effectiveness of our design choices, we carry out an ablation study on BSD100 using HiT-SIR as the baseline and compare four model variants.

Table \ref{table3} summarizes the results.
The notation “$\ast$3” in the baseline parameter column indicates that HiT-SIR uses separate models for each scale; in such cases, the score reported at each scale corresponds to the best among the respective pretrained models. For our variants, we use the notation Norm-Experts, where BN refers to Batch Normalization, SA to SimAM, and the following number (e.g., 4, 16, 64) denotes the number of expert weights in the SAAM block.
Replacing BatchNorm with SimAM (SA-4) leads to consistent improvements in PSNR and SSIM across all scales, with negligible increase in parameter count. Increasing the expert count to 16 (Ours) further improves performance, achieving the best accuracy-to-size trade-off at 903K parameters. While removing the dense layer entirely (SA-16 no-dense) achieves the highest scores in Table~\ref{table3}, it introduces a 93K parameter overhead, which is significant for extremely compact architectures like SCNet (~150K). Notably, retaining the dense layer yields comparable accuracy with substantially fewer parameters.
Expanding to 64 experts (SA-64) does not provide extra benefit and even slightly reduces the performance, indicating that the configuration with 16 experts and a lightweight dense layer offers the most efficient balance between performance and model size. Accordingly, our plug-in is designed to optimize this trade-off for practical deployment in lightweight SR architectures.

\section{Conclusion}
\label{sec:Conclusion}
In this work, we introduce \textbf{SAAM}, a lightweight plug-in module that equips fixed-scale super-resolution models with arbitrary-scaling. SAAM combines SimAM attention, efficient scale-aware convolutions, and gradient variance loss to deliver high-quality and consistent results across a wide range of scaling factors—all while adding minimal computational overhead. Comprehensive experiments and ablation studies across various SR architectures demonstrate that SAAM not only generalizes well to unseen and non-integer scales but also outperforms existing approaches in both accuracy and efficiency. These findings highlight SAAM’s practicality and versatility for real-world SR tasks that demand adaptable resolution enhancement.
\clearpage
\section*{Acknowledgement}
\label{sec:Acknowledgement}
This work was supported by Institute of Information \& communications Technology Planning \& Evaluation (IITP) grant funded by the Korea government(MSIT) (RS-2025-02215122, Development and Demonstration of Lightweight AI Model for Smart Homes)

\small
{
    \small
    \bibliographystyle{ieeenat_fullname}
    \bibliography{main}
}

\clearpage
\appendix

\section{Appendix}

\begin{figure*}[!b] 
  \centering
  \includegraphics[scale=0.625]{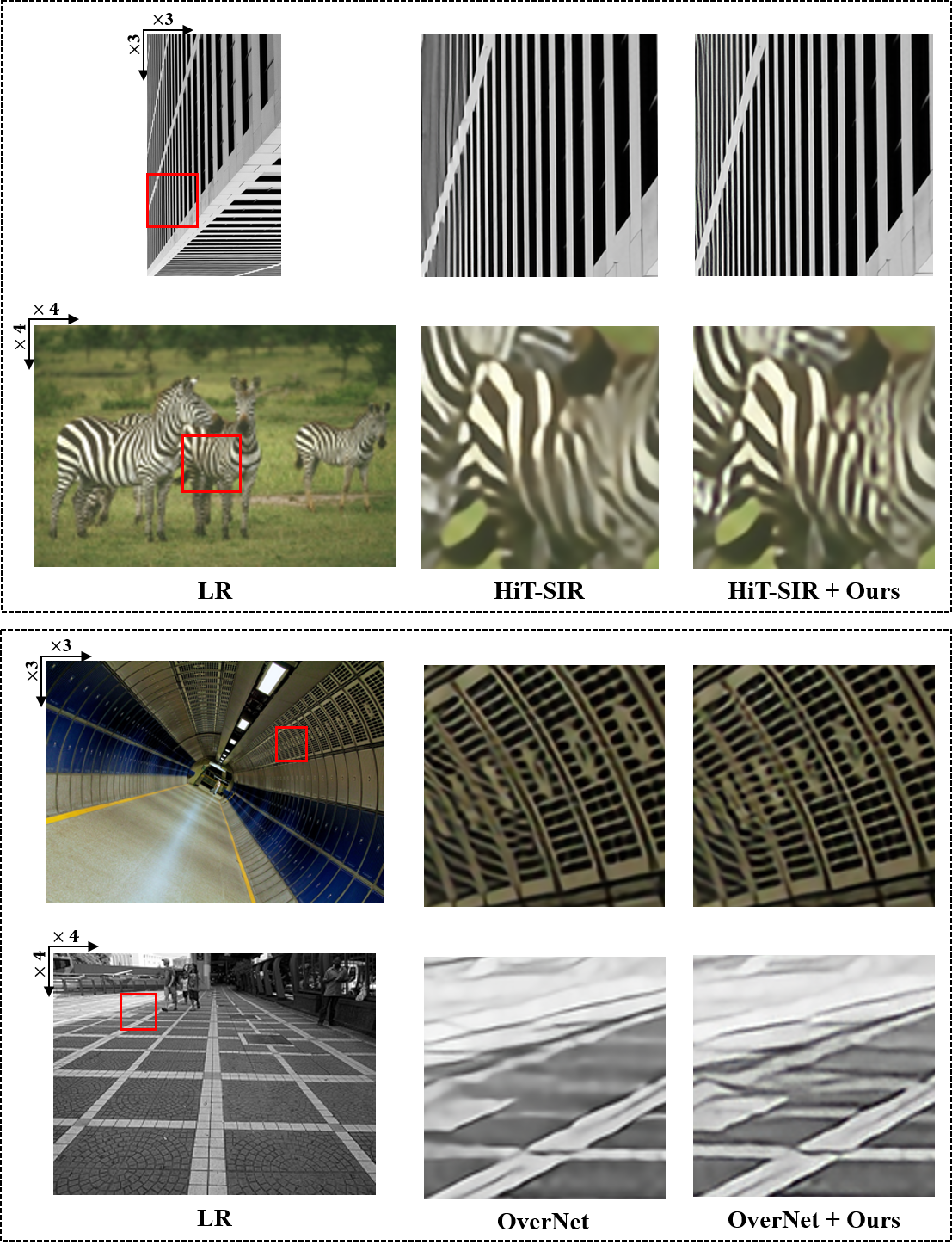}
  \caption{Qualitative comparison by integrating our method into the baseline model on BSD100 and Urban100.}
  \label{fig:fig3}
\end{figure*}

This section provides supplementary materials related to our SAAM plug-in, including the following components:

\begin{itemize}
\setlength{\itemsep}{0pt}
\setlength{\parsep}{0pt}
    \item Additional quantitative comparisons on another benchmark datasets (Set5~\cite{set5} and Set14~\cite{set14}).
    \item Qualitative comparison between the baseline and our plug-in-integrated model on BSD100~\cite{b100} and Urban100~\cite{u100}.
\end{itemize}

\subsection{Qualitative Comparisions}
Figure~\ref{fig:fig3} illustrates that our method achieves clearer and more compelling results than the existing approaches~\cite{hitsr, overnet}.
Especially, existing methods sometimes produce blurred images or distort structural details during the super-resolution process, whereas our proposed method preserves both sharpness and shape.

\subsection{Additional Quantitative Comparisons}
As shown in Table~\ref{table4}, the existing methods~\cite{scnet, hitsr, overnet} were trained solely on datasets matching their specific output scales.
Even though the number of dataset samples is small, our proposed method demonstrates robust multi-scale handling capabilities, often outperforming counterparts trained on single scales.

\begin{table*}[!t]
\centering
\resizebox{\textwidth}{!}{
\begin{tabular}{|lcccccc}
\hline
\multicolumn{1}{|l||}{Scale}                                                             & \multicolumn{2}{c||}{×2}                                                   & \multicolumn{2}{c||}{×3}                                                   & \multicolumn{2}{c||}{×4}                                                   \\ \hline
\multicolumn{1}{|l||}{Metrics}                                                           & \multicolumn{2}{c||}{PSNR / SSIM}                                          & \multicolumn{2}{c||}{PSNR / SSIM}                                          & \multicolumn{2}{c||}{PSNR / SSIM}                                          \\ \hline
\multicolumn{1}{|l||}{Dataset}                                                           & \multicolumn{1}{c|}{Set5}           & \multicolumn{1}{c||}{Set14}          & \multicolumn{1}{c|}{Set5}           & \multicolumn{1}{c||}{Set14}          & \multicolumn{1}{c|}{Set5}           & \multicolumn{1}{c||}{Set14}          \\ \hline\hline
\multicolumn{1}{|l||}{\begin{tabular}[c]{@{}l@{}}SCNet x2 \\ (172K)\end{tabular}}        & \multicolumn{1}{l|}{37.65 / 0.9593} & \multicolumn{1}{l||}{33.30 / 0.9143} & \multicolumn{1}{c|}{-}              & \multicolumn{1}{c||}{-}              & \multicolumn{1}{c|}{-}              & \multicolumn{1}{c||}{-}              \\ \hline
\multicolumn{1}{|l||}{\begin{tabular}[c]{@{}l@{}}SCNet x3 \\ (172K)\end{tabular}}        & \multicolumn{1}{c|}{-}              & \multicolumn{1}{c||}{-}              & \multicolumn{1}{l|}{33.89 / 0.9233} & \multicolumn{1}{l||}{30.00 / 0.8350} & \multicolumn{1}{c|}{-}              & \multicolumn{1}{c||}{-}              \\ \hline
\multicolumn{1}{|l||}{\begin{tabular}[c]{@{}l@{}}SCNet x4 \\ (172K)\end{tabular}}        & \multicolumn{1}{c|}{-}              & \multicolumn{1}{c||}{-}              & \multicolumn{1}{c|}{-}              & \multicolumn{1}{c||}{-}              & \multicolumn{1}{l|}{31.70 / 0.8887} & \multicolumn{1}{l||}{28.32 / 0.7746} \\ \hline
\multicolumn{1}{|l||}{\begin{tabular}[c]{@{}l@{}}SCNet+Ours(T)\\ (901K)\end{tabular}}    & \multicolumn{1}{l|}{\textbf{37.72 / 0.9595}} & \multicolumn{1}{l||}{\textbf{33.31 / 0.9152}} & \multicolumn{1}{l|}{33.82 / 0.9213} & \multicolumn{1}{l||}{29.93 / 0.8341} & \multicolumn{1}{l|}{30.92 / 0.8740} & \multicolumn{1}{l||}{27.83 / 0.7588} \\ \hline
\multicolumn{1}{|l||}{\begin{tabular}[c]{@{}l@{}}SCNet+Ours(L)\\ (1.38M)\end{tabular}}   & \multicolumn{1}{l|}{\textbf{37.65 / 0.9596}} & \multicolumn{1}{l||}{33.26 / \textbf{0.9158}} & \multicolumn{1}{l|}{\textbf{34.08 / 0.9249}} & \multicolumn{1}{l||}{\textbf{30.02 / 0.8387}} & \multicolumn{1}{l|}{31.51 / 0.8840} & \multicolumn{1}{l||}{28.14 / 0.7689} \\ \hline\hline
\multicolumn{7}{|l|}{}                                                                                                                                                                                                                                                                                                      \\ \hline
\multicolumn{1}{|l||}{\begin{tabular}[c]{@{}l@{}}HiT-SIR x2 \\ (772K)\end{tabular}}      & \multicolumn{1}{c|}{38.22 / 0.9613} & \multicolumn{1}{c||}{33.91 / 0.9213} & \multicolumn{1}{c|}{-}              & \multicolumn{1}{c||}{-}              & \multicolumn{1}{c|}{-}              & \multicolumn{1}{c||}{-}              \\ \hline
\multicolumn{1}{|l||}{\begin{tabular}[c]{@{}l@{}}HiT-SIR x3\\ (772K)\end{tabular}}       & \multicolumn{1}{c|}{-}              & \multicolumn{1}{c||}{-}              & \multicolumn{1}{c|}{34.72 / 0.9298} & \multicolumn{1}{c||}{30.62 / 0.8474} & \multicolumn{1}{c|}{-}              & \multicolumn{1}{c||}{-}              \\ \hline
\multicolumn{1}{|l||}{\begin{tabular}[c]{@{}l@{}}HiT-SIR x4\\ (772K)\end{tabular}}       & \multicolumn{1}{c|}{-}              & \multicolumn{1}{c||}{-}              & \multicolumn{1}{c|}{-}              & \multicolumn{1}{c||}{-}              & \multicolumn{1}{c|}{32.51 / 0.8991} & \multicolumn{1}{c||}{28.84 / 0.7873} \\ \hline
\multicolumn{1}{|l||}{\begin{tabular}[c]{@{}l@{}}HiT-SIR+Ours(T)\\ (901K)\end{tabular}}  & \multicolumn{1}{c|}{38.21 / \textbf{0.9614}} & \multicolumn{1}{c||}{\textbf{34.06 / 0.9223}} & \multicolumn{1}{c|}{34.69 / \textbf{0.9298}} & \multicolumn{1}{c||}{30.56 / 0.8485} & \multicolumn{1}{c|}{32.47 / 0.8986} & \multicolumn{1}{c||}{28.78 / \textbf{0.7893}} \\ \hline
\multicolumn{1}{|l||}{\begin{tabular}[c]{@{}l@{}}HiT-SIR+Ours(L)\\ (1.38M)\end{tabular}} & \multicolumn{1}{c|}{\textbf{38.27 / 0.9614}} & \multicolumn{1}{c||}{\textbf{33.95 / 0.9223}} & \multicolumn{1}{c|}{\textbf{34.74 / 0.9299}} & \multicolumn{1}{c||}{30.55 / \textbf{0.8482}} & \multicolumn{1}{c|}{\textbf{32.52 / 0.8991}} & \multicolumn{1}{c||}{28.77 / \textbf{0.7888}} \\ \hline\hline
\multicolumn{7}{|l}{}                                                                                                                                                                                                                                                                                                       \\ \hline
\multicolumn{1}{|l||}{\begin{tabular}[c]{@{}l@{}}OverNet x2 \\ (0.9M)\end{tabular}}      & \multicolumn{1}{c|}{38.11 / 0.9607} & \multicolumn{1}{c||}{33.71 / 0.9183} & \multicolumn{1}{c|}{32.06 / 0.8912} & \multicolumn{1}{c||}{28.89 / 0.8083} & \multicolumn{1}{c|}{30.17 / 0.8437} & \multicolumn{1}{c||}{27.36 / 0.7393} \\ \hline
\multicolumn{1}{|l||}{\begin{tabular}[c]{@{}l@{}}OverNet x3 \\ (0.9M)\end{tabular}}      & \multicolumn{1}{c|}{37.44 / 0.9582} & \multicolumn{1}{c||}{33.29 / 0.9157} & \multicolumn{1}{c|}{34.46 / 0.9273} & \multicolumn{1}{c||}{30.41 / 0.8429} & \multicolumn{1}{c|}{31.97 / 0.8905} & \multicolumn{1}{c||}{28.49 / 0.7770} \\ \hline
\multicolumn{1}{|l||}{\begin{tabular}[c]{@{}l@{}}OverNet x4 \\ (0.9M)\end{tabular}}      & \multicolumn{1}{c|}{36.25 / 0.9525} & \multicolumn{1}{c||}{32.41 / 0.9070} & \multicolumn{1}{c|}{33.89 / 0.9228} & \multicolumn{1}{c||}{30.08 / 0.8375} & \multicolumn{1}{c|}{32.26 / 0.8958} & \multicolumn{1}{c||}{28.64 / 0.7821} \\ \hline
\multicolumn{1}{|l||}{\begin{tabular}[c]{@{}l@{}}OverNet+Ours(T)\\ (1.07M)\end{tabular}} & \multicolumn{1}{l|}{38.01 / 0.9606} & \multicolumn{1}{l||}{33.66 / \textbf{0.9186}} & \multicolumn{1}{l|}{\textbf{34.46 / 0.9277}} & \multicolumn{1}{l||}{30.30 / \textbf{0.8436}} & \multicolumn{1}{l|}{32.21 / 0.8957} & \multicolumn{1}{l||}{28.53 / \textbf{0.7841}} \\ \hline
\multicolumn{1}{|l||}{\begin{tabular}[c]{@{}l@{}}OverNet+Ours(L)\\ (2.3M)\end{tabular}}  & \multicolumn{1}{c|}{\textbf{38.13 / 0.9609}} & \multicolumn{1}{c||}{\textbf{33.87 / 0.9194}} & \multicolumn{1}{c|}{\textbf{34.54 / 0.9284}} & \multicolumn{1}{c||}{30.34 / \textbf{0.8446}} & \multicolumn{1}{c|}{32.24 / \textbf{0.8965}} & \multicolumn{1}{c||}{\textbf{28.58 / 0.7855}} \\ \hline
\end{tabular}
}
\caption{Additional quantitative comparisons by integrating our method into existing models. \textbf{Bold} indicates superior performance over baseline.}
\label{table4}
\end{table*}

\end{document}